\documentclass{article}
\usepackage{ijcai21}

\usepackage[linesnumbered,ruled,vlined]{algorithm2e}
\usepackage{times}
\usepackage{soul}
\usepackage{url}
\usepackage{color}
\usepackage[hidelinks]{hyperref}
\usepackage[utf8]{inputenc}
\usepackage[small]{caption}
\usepackage{graphicx}
\usepackage{amsmath}
\usepackage{amsthm}
\usepackage{booktabs}
\usepackage{multirow}
\usepackage{epsfig}
\usepackage{amssymb}
\usepackage{mathtools}
\usepackage{multirow}
\usepackage{subcaption}
\usepackage{xcolor}
\usepackage{xspace}

\newcommand{\eat}[1]{}

\newcommand{\norm}[1]{\left\lVert#1\right\rVert}
\newcommand{\fixnorm}[1]{\lVert#1\rVert}

\DeclareMathOperator*{\argmin}{arg\,min}

\newcommand{\vect}[1]{\boldsymbol{#1}}

\makeatletter
\DeclareRobustCommand\onedot{\futurelet\@let@token\@onedot}
\def\@onedot{\ifx\@let@token.\else.\null\fi\xspace}

\def\eg{\emph{e.g}\onedot} 
\def\ie{\emph{i.e}\onedot} 
 
\def\etc{\emph{etc}\onedot} 
\def\wrt{w.r.t\onedot}

\mathchardef\mhyphen="2D 
\makeatother

\title{Feature Space Targeted Attacks by Statistic Alignment}

\author{
Lianli Gao\and
Yaya Cheng\and
Qilong Zhang\and
Xing Xu\And
Jingkuan Song\thanks{corresponding author}
\affiliations
Center for Future Media, University of Electronic Science and Technology of China
\emails
yaya.cheng@hotmail.com,
qilong.zhang@std.uestc.edu.cn,
jingkuan.song@gmail.com
}

\begin{document}

\maketitle

\begin{abstract}
By adding human-imperceptible perturbations to images, DNNs can be easily fooled. As one of the mainstream methods, feature space targeted attacks perturb images by modulating their intermediate feature maps, for the discrepancy between the intermediate source and target features is minimized. However, the current choice of pixel-wise Euclidean Distance to measure the discrepancy is questionable because it unreasonably imposes a spatial-consistency constraint on the source and target features.
Intuitively, an image can be categorized as ``cat'' no matter the cat is on the left or right of the image. 
To address this issue, we propose to measure this discrepancy using statistic alignment. Specifically, we design two novel approaches called Pair-wise Alignment Attack and Global-wise Alignment Attack, which attempt to measure similarities between feature maps by high-order statistics with translation invariance. Furthermore, we systematically analyze the layer-wise transferability with varied difficulties to obtain highly reliable attacks. Extensive experiments verify the effectiveness of our proposed method, and it outperforms the state-of-the-art algorithms by a large margin. Our code is publicly available at \url{https://github.com/yaya-cheng/PAA-GAA}.
\end{abstract}

\section{Introduction}
\label{introduction}
Deep neural networks (DNNs)~\cite{resnet,dense,vgg,inc-v3} have made impressive achievements in these years, and various fields are dominated by them, \eg, object detection~\cite{yolo}. However, recent works demonstrate that DNNs are highly vulnerable to the adversarial examples~\cite{fgsm,evasion} which are only added with human-imperceptible perturbations. To find out the insecure ``bugs" in the DNNs, many works pay attention to the generation of adversarial examples. 

In general, the attack methods can be grouped into three broad categories: white-box, gray-box, and black-box attacks. For the white-box setting~\cite{deepfool,c-w}, the adversaries can access all information (\eg, the architectures and parameters) of the victim's models. Thus the update directions of the adversarial examples are accurate. For the gray-box setting~\cite{nes,Bayes}, only the output logits or labels are available. Therefore, most of the works craft adversarial examples through a considerable amount of queries. However, in many scenarios, both the white-box and the gray-box attacks are infeasible owing to the opaque deployed models. For the black-box setting, all information of the victim's models is unavailable. Since the decision boundaries of different DNNs are similar, the resultant adversarial examples crafted for the substitute models, \eg, well-trained models, are also practical for others, which is called the transferability of adversarial examples. Most black-box attack methods~\cite{mifgsm,feature,pifgsm,pifgsm++,NISI} aim at enhancing the transferability of adversarial examples depending on information from the classification layers of the substitute models. However, it is still challenging to improve the success rate of black-box targeted attacks, \ie, induce the victim's models to predict the pre-set target labels.

To tackle the poor effectiveness of black-box targeted attacks, researchers~\cite{feature16,feature} delve into the feature space targeted attacks, which perturb images by modulating their intermediate feature maps.
For example, given a source image,~\cite{feature} first select a single sample of the target label whose intermediate activation is furthest from the source one under Euclidean distance. Then, perturbation is crafted by minimizing the Euclidean distance between the source and target features.
However, since Euclidean distance prefers to focus on the spatial gap between two features, it will select the spatially furthest target image rather than the one with the outermost semantic meaning. For instance, considering a source image with a cat on the left and the target label is ``dog", under the above setting, the algorithm tends to choose a target image that has a dog on the right instead of on the left. When it comes to the generation of perturbation, what the algorithm needs to do is the semantic meaning alignment between the source and target features and the minimization of the huge spatial discrepancy. 
Overall, the current choice of pixel-wise Euclidean distance to measure the discrepancy is questionable, as it unreasonably imposes a spatial-consistency constraint on the source and target features.

To produce spatial-agnostic measurements, we propose two novel approaches called Pair-wise Alignment Attack and Global-wise Alignment Attack, which attempt to measure similarities between features by high-order statistics with translation invariance. 
With this perspective, we deal with the feature space targeted attacks as the problem of statistic alignment. By aligning the source and target high-order statistics, rather than depending on the Euclidean distance, we can make the two feature maps semantically close without introducing an excessive spatial gap in feature space. 

To sum up, our contribution is summarized as three-folds:
1) We point out that the current choice of pixel-wise Euclidean Distance to measure the discrepancy between two features is questionable, for it unreasonably imposes a spatial-consistency constraint on the source and target features. By exploring high-order statistics with translation invariance, two novel methods are proposed: a) Pair-wise Alignment Attack and b) Global-wise Alignment Attack, which deal with feature space targeted attacks as a problem of statistic alignment;
2) To obtain high-reliability results, we systematically analyze the layer-wise transferability. Furthermore, to set all images under the same transfer difficulty, which ranges from the easiest to the hardest, we assign the target labels of the same difficulty level to them and give a comprehensive evaluation of our methods.
and 
3) Extensive experimental results show the effectiveness of our methods, which outperform the state-of-the-art by $6.92\%$ at most and $1.70\%$ on average in typical setups.

\section{Related Works}
After the discovery of adversarial examples~\cite{fgsm,evasion}, many excellent works are proposed. Generally, based on different goals, attack methods can be divided into non-targeted attacks and targeted attacks. For non-targeted attacks (\eg,~\cite{difgsm}), all need to do is fooling DNNs to misclassify the perturbed images. For targeted attacks, the adversaries must let the DNNs predict specific untrue labels for the adversarial examples.~\cite{pom} apply Poincar$\acute{e}$ distance and Triplet loss to regularize the targeted attack process.
~\cite{ssm} propose staircase sign method to utilize the gradients of the substitute models effectively.
The above methods craft adversarial examples by directly using the outputs of the classification layers, \ie, logits (un-normalized log probability).

In addition to these, researchers~\cite{YosinskiCBL14} observe that distorting the features in the intermediate layers of DNNs can also generate transferable adversarial examples. Based on this,~\cite{feature} generate adversarial examples by minimizing the Euclidean distance between the source and target feature maps.~\cite{F} leverage class-wise and layer-wise deep feature distributions of substitute models
.~\cite{F_E} extract feature hierarchy of DNNs to boost the performance of targeted adversarial attacks further. However, the above methods need to train specific auxiliary classifiers for each target label, thus suffering from expensive computation costs. 

\eat{\subsection{Domain Transfer}
As the dataset bias~\cite{biased} and covariate shift~\cite{covariate} are found, the generalization abilities of models that measure how well they can perform in other datasets and tasks gain increasingly attention~\cite{beijbom2012domain,patel2015visual,chen2012marginalized,saenko2010adapting,gong2012geodesic}. One of the key components in domain transfer is bridging the gap between the source and target distribution. In~\cite{twoSampleTest}, they define the \textit{two-sample problem}. Through matching \textit{embedding mean} in a RKHS(Reproducing Kernel Hilbert Space), they propose the Maximum Mean Discrepancy(MMD) statistic for \textit{two-sample} testing. There are many methods for statistic alignment, such as non-negative embedding~\cite{redko2016non}, sample selections~\cite{huang2006novel,li2016sample}, principal axes alignment~\cite{aljundi2015landmarks,fernando2013unsupervised,fernando2015joint}. Different from the previous methods, where additional components are required.~\cite{bn} find that when DNNs architecture goes deeper, the two order moment(\textit{mean} and \textit{variance}) of Batch Normalization layers contain more complicate traits to represent different distribution. With this perspective, they apply a simple but effective method by utilizing these traits.
}

\section{Methodology}
In this section, we first give some notations of targeted attacks, and the untargeted version can be simply derived. Then we describe our proposed methods, i.e., Pair-wise Alignment Attack and Global-wise Alignment Attack, in Subsection \ref{DTAMMD} and \ref{sec.global}. The attack process is detailed in Subsection \ref{sec.alg}.

\subsection{Preliminaries}

\paragraph{Adversarial targeted attacks.} 
This task aims at \textit{fooling} a DNN $\mathcal{F}$ to misclassify perturbed image $\vect{x^{adv}} = \vect{x} + \vect\delta$, where $\vect{x}$ is the original image of label $y$, $\vect{ \delta}$ is an imperceptible perturbation added on $\vect x$. In our work, $\ell_\infty$-norm is applied to evaluate the imperceptibility of perturbation, i.e., $\norm{\vect{\delta}}_\infty \leq \epsilon$. Different from the untargeted attacks that only need to let $\mathcal{F}$ will not perform correct recognition, targeted attacks restrict the misclassified label to be $y^{tgt}$. The constrained optimization of targeted attacks can be written as:
\begin{equation}
        \vect{x^{adv}}=\argmin \mathcal{L}(\vect{x^{adv}}, y^{tgt}),
        \mathit{s.t.} \norm{\vect{x^{adv}}-\vect x}_\infty \leq \epsilon,
\end{equation}
where $\mathcal{L(\cdot,\cdot)}$ is the loss function to calculate perturbations.

\paragraph{Perceptions of DNNs.} DNNs, especially convolutional neural networks (CNNs), have their patterns to perceive and understand images~\cite{VUVCN}, which is caused by the mechanism of convolutional layers. As introduced in \cite{translationInvariance}, convolution kernels do not perform a one-time transformation to produce result from the input. Instead, a small region of input is perceived iteratively so that features at every layer still hold local structures similar to that of the input (see Appendix~\ref{appendix.perception}). This property of convolution leads to the translation homogeneity of intermediate feature maps. Therefore, measuring only the Euclidean distance between two feature maps will be inaccurate when there are translations, rotations, \etc.

\subsection{Pair-wise Alignment Attack}
\label{DTAMMD}
Given an image $\vect{x}^{tgt}$ of target label $y^{tgt}$, a specific intermediate layer $l$ from network $\mathcal{F}$. We use $\vect{S}^l\in\mathbb{R}^{{N_l}\times{M_l}}$ to denote the feature of $\vect{x^{adv}}$ at layer $l$ of $\mathcal{F}$. Similarly, $\vect{T}^l\in\mathbb{R}^{{N_l}\times{M_l}}$ is the feature of $\vect{x}^{tgt}$. Specifically, $N_l$ is the number of channels and $M_l$ is the product of the height and width of features. 

As described before, since Euclidean distance imposes unreasonable spatial-consistency constraint on $\vect{S}^l$ and $\vect{T}^l$, choosing it as the metric leads to redundant efforts on spatial information matching. To handle this, we propose the Pair-wise Alignment Attack ($\mathbf{PAA}$). Assuming that the label information is modeled by highly abstract features, we denote $\vect{S}^l$ and $\vect{T}^l$ are under two distributions $p$ and $q$, which models the label information $y$ and $y^{tgt}$, respectively. Naturally, an arbitrary feature extracted from $\mathcal{F}$ is treated as a sample set of a series of feature vectors over corresponding distribution. 

So the problem is how to utilize these samples to further estimate the difference between $p$ and $q$. Empirically, source and target sample sets $\Omega \sim p$, $Z \sim q$ are built by splitting $\vect{S}^l$, $\vect{T}^l$ into individual vectors, where $\Omega\!=\!\{{\vect s_{\cdot i}}\}_{i=1}^{M_l}$,  $Z\!=\!\{{\vect t_{\cdot j}}\}_{j=1}^{M_l}$. Another way of splitting in where $\Omega\!=\!\{{\vect s_{ i\cdot}}\}_{i=1}^{N_l}$,  $Z\!=\!\{{\vect t_{j\cdot}}\}_{j=1}^{N_l}$ is analysed in Appendix~\ref{appendix.sampleStrategy} After that, through measuring the similarity of $\Omega$ and $Z$, the discrepancy between $p$ and $q$ is estimated. Typically, this is a two-sample problem~\cite{twoSampleTest}. 


As introduced in~\cite{twoSampleTest}, $\operatorname{MMD}^{2}$ has been explored for the two-sample problem. Let $\mathcal{H}$ be a reproducing kernel Hilbert space (RKHS) with an associated continuous kernel $k(\cdot,\cdot)$. For all  $f\!\in\!\mathcal{H}$, the \textit{mean embedding} of $p$ in $\mathcal{H}$ is an unique element $\mu_{p}$ which satisfies the condition of ${\mathbb{E}_{\omega\sim{p}}f\!=\!\langle f,\mu_{p}\rangle}_\mathcal{H}$. Then in our task, $\operatorname{MMD}^{2}[p, q]$ is defined as the RKHS distance between $\mu_{p}$ and $\mu_{q}$:
\begin{equation}
    \label{Eq.MMD}
    \begin{split}
        &\operatorname{MMD}^{2}[p, q] 
        =~~ \fixnorm{\mu_{p}-\mu_{q}}^2_{\mathcal{H}} \\[3.5pt]
        =&\; {\langle \mu_{p},\mu_{p}\rangle}_{\mathcal{H}}+{\langle \mu_{q},\mu_{q}\rangle}_{\mathcal{H}}-2{\langle \mu_{p},\mu_{q}\rangle}_{\mathcal{H}}\\[1pt]
        =&\frac{1}{M_l^2}\sum\limits_{i,j = 1}^{M_l}{k({\vect{s}_{\cdot i}},\vect{s}_{\cdot j}})+\frac{1}{M_l^2}\sum\limits_{i,j = 1}^{M_l} {k(\vect{t}_{\cdot i},\vect{t}_{\cdot j}})\\[1pt]
        &- \frac{2}{M_l^2}\sum\limits_{i,j = 1}^{M_l,M_l}{k(\vect{s}_{\cdot i},\vect{t}_{\cdot j}}).
    \end{split}
\end{equation}

Specifically, $\operatorname{MMD}^{2}$ is calculated by two kinds of pairs: a) intra-distribution pairs $\left(\vect{s}_{\cdot i}, \vect{s}_{\cdot j}\right)$, $\left(\vect{t}_{\cdot i}, \vect{t}_{\cdot j}\right)$ and b) inter-distribution pair $\left(\vect{s}_{\cdot i}, \vect{t}_{\cdot j}\right)$. Obviously, $\operatorname{MMD}^{2}$ is not affected by spatial translations, \ie, shifting or rotation will not change the result of equation \ref{Eq.MMD}, which is the key difference from Euclidean distance. Furthermore, based on the critical property $\operatorname{MMD}^{2}[p, q]\!=\!0\,\mathit{iff}\, p\!=\!q$~\cite{twoSampleTest}, minimizing equation \ref{Eq.MMD} equals to modulating source feature to target's:
\begin{equation}
\label{Eq.lossmmd}
    \begin{split}
        \mathcal{L_P}{(\vect{S}^l, \vect{T}^l)}=&\operatorname{MMD}^{2}[p, q].
    \end{split}
\end{equation}
Since the kernel choice plays a key role in the mean embedding matching~\cite{twoSampleTest}. In our experiments, three kernel functions will be studied to evaluate their effectiveness in statistic alignment:
\begin{itemize}
  \item Linear kernel $\mathbf{PAA_\ell}$: $k(\vect{s},\vect{t})\!=\!{\vect{s}^{\scriptscriptstyle T}}\vect{t}$.
  \item Polynomial kernel $\mathbf{PAA_p}$: $k(\vect{s},\vect{t})\!=\!{(\vect{s}^{\scriptscriptstyle T}\vect{t}+c)}^d$.
  \item Gaussian kernel $\mathbf{PAA_g}$: $k(\vect s,\vect t)\!=\!\exp {(-{\fixnorm{\vect s - \vect t}_2^2 \over {2\sigma^2}})}$,
\end{itemize}
where bias $c$, power $d$ and variance $\sigma^2$ are hyper-parameters. Following~\cite{feature}, by randomly sampling images from each label, a gallery is maintained for picking target images. With the help of the gallery, the pipeline of getting $\vect{x}^{tgt}$ by $\mathbf{PAA}$ is as follows: Given a source image $\vect x$, we obtain $y^{tgt}$ by using different strategies of target label selection. After that, $\vect{x}^{tgt}$ is chosen from the corresponding sub-gallery by finding the image with the largest loss $\mathcal{L_P}$. It is worth noting that we adopt the linear-time unbiased estimation of $\operatorname{MMD}^{2}[p, q]$ from~\cite{twoSampleTest} to decrease the space and computation complexity during the selection of the target image $\vect{x}^{tgt}$.

\subsection{Global-wise Alignment Attack}
\label{sec.global}
Since Pair-wise Alignment Attack involves time-consuming pair-wise computation, we propose the other efficient approach that achieves comparable performance. Unlike the previous one, Global-wise Alignment Attack (\textbf{GAA}) explicitly matches moments of source, and target sample sets $\Omega$, $Z$. 
Specifically, we employ two global statistics: a) first-order raw moment (\textit{mean}) and b) second-order central moment (\textit{variance}) to guide the modulation of features. Let $\mu_{\vect {S}^l}^i$, $\mu _{\vect {T}^l}^i$, $\sigma _{\vect {S}^l}^i$, $\sigma _{\vect {T}^l}^i$ be the mean and variance of the $i$th channel of $\vect S^l$ and $\vect T^l$, respectively: 
\begin{align}
    &\mu _{\vect{S}^l}^i = \frac{1}{M_l}\sum\nolimits_{j = 1}^{M_l} {(\vect{S}^l)}_{ij},\; {\sigma_{\vect{S}^l}^i} = \operatorname{Var}((\vect{S}^l)_{i \cdot}), \\
    &\mu _{\vect{T}^l}^i = \frac{1}{M_l}\sum\nolimits_{j = 1}^{M_l} {(\vect{T}^l)_{ij}}, \;{\sigma_{\vect{T}^l}^i} = \operatorname{Var}({(\vect{T}^l)}_{i \cdot}).
\end{align}
Minimizing the gaps between $\Omega$ and $Z$ of these two moments equals to aligning the source and target features globally:
\begin{equation}
\label{Eq.losss}
    \begin{gathered}
        \delta_\mu = \norm{\mu_{\vect{S}^l}^i - \mu_{\vect{T}^l}^i}, 
        \delta_\sigma = \norm{\sigma_{\vect{S}^l}^i - \sigma_{\vect{T}^l}^i}, \\[3.5pt]
        \mathcal{L_G}{(\vect{S}^l, \vect{T}^l)} = \delta_\mu + \delta_\sigma.
    \end{gathered}
\end{equation}

The reasons for performing Global-wise Alignment are: 1) the two moments are practical to estimate the distribution on a dataset, just like what batch-normalization does; and 2) when the architectures of DNNs go deeper, these two moments will contain more complicated traits to represent different distributions~\cite{bn}. Similar as $\mathbf{PAA}$, $\mathbf{GAA}$ also chooses the target image from the gallery by calculating Equation~(\ref{Eq.losss}).

\begin{figure}[t]
    \centering
    \includegraphics[width=0.85\columnwidth,height=7.5cm]{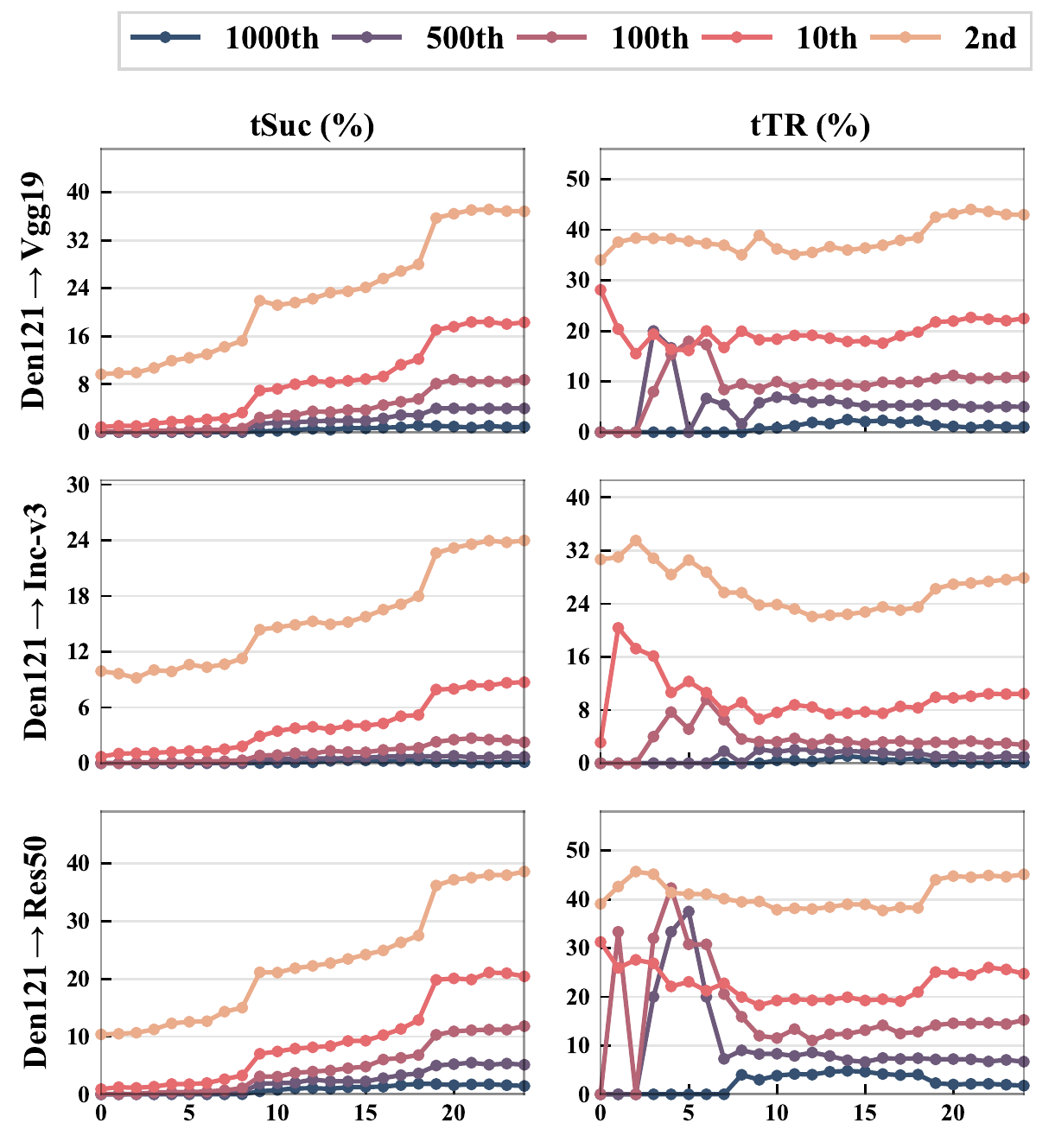}
    \caption{Performance (tSuc and tTR) of $\mathbf{PAA_p}$ \wrt $\mathit{2nd}$, $10th$, $100th$, $500th$, and $1000th$ settings. Target label of higher ranking leads to better performance.}
    \label{Fig.classPAA}
\end{figure}

\subsection{Attack Algorithm}
\label{sec.alg}
Motivated by MIFGSM~\cite{mifgsm} which using momentum to memorize previous gradients and follow the setting of AA~\cite{feature}, we integrate momentum to the pipeline of perturbation generation. Specifically, for two kinds of attacks, \ie, \textbf{PAA} and \textbf{GAA}, we firstly calculate gradients step-by-step:
\begin{equation}
    \vect{g}_\nu = {\nabla _{\vect{x}^{adv}_{_\nu }}} \mathcal{L}(\vect{S}^l_\nu,\vect{T}^l),
\end{equation}
where $\nu$ is the current step during the whole iteration, $\vect{S}_{\nu}^l$ is the intermediate feature of the perturbed image $\vect{x}^{adv}_\nu$ at iteration $\nu$, and $\vect{x}_0^{adv}\!=\! \vect{x}$. Then the momentum term is accumulated by previous gradients:
\begin{equation}
    \begin{split}
        {\beta _{\nu  + 1}} = {\mu \cdot \beta _\nu } + \frac{\vect{g}_\nu}{{\left\| {\vect{g}_\nu} \right\|}},
    \end{split}
\end{equation}
where $\mu$ refers to the decay factor, ${\beta_ \nu}$ is the momentum term at iteration $\nu$ and ${\beta_ 0}$ is initialized to 0. Finally, under the $\ell_\infty$-norm constraint, adversarial examples are crafted by performing the above calculations iteratively:
\begin{equation}
    \begin{split}
        {\vect{x}_{\nu + 1}^{adv}} = \operatorname{clip}_{\vect{x},\epsilon}({\vect{x}^{adv}_\nu} - \alpha \cdot \operatorname{sign}({\beta _{\nu+1} })), 
    \end{split}    
\end{equation}
where $\alpha$ is a given step size.

\begin{table}[t]
    \resizebox{\columnwidth}{!}{%
    \begin{tabular}{@{}ccccccc@{}}
    \toprule
    \multirow{2}{*}{} & \multicolumn{2}{c}{Den121$\rightarrow$VGG19} & \multicolumn{2}{c}{Den121$\rightarrow$Inc-v3} & \multicolumn{2}{c}{Den121$\rightarrow$Res50} \\
                             & tSuc             & tTR            & tSuc             & tTR            & tSuc             & tTR             \\ \midrule\midrule
    TIFGSM                  &$0.40$ &$0.41$ &$0.08$ &$0.08$ &$0.24$ &$0.24$ \\
    MIFGSM                  &$1.48$ &$1.50$ &$0.54$ &$0.55$ &$2.44$ &$2.46$ \\
    AA                      & $1.18$          & $1.61$        & $0.50$          & $0.68$        & $1.78$           & $2.32$         \\
    $\mathbf{GAA}$          & $3.20$          & $4.17$ & $0.70$          & $0.91$        & $4.22$          & $5.62$         \\
    $\mathbf{PAA_g}$        & $1.60$          & $2.52$        & $0.52$          & $0.73$        & $2.42$          & $3.60$         \\
    $\mathbf{PAA_\ell}$     & $3.20$ & $3.97$ & $0.74$ & $0.94$ & $4.40$ & $5.65$ \\
    $\mathbf{PAA_{p}}$      & $\mathbf{4.38}$ &$\mathbf{5.56}$& $\mathbf{1.16}$ &$\mathbf{1.45}$& $\mathbf{6.08}$ & $\mathbf{7.95}$\\\midrule\midrule
    
    \multirow{2}{*}{} & \multicolumn{2}{c}{VGG19$\rightarrow$Inc-v3} & \multicolumn{2}{c}{VGG19$\rightarrow$Den121} & \multicolumn{2}{c}{VGG19$\rightarrow$Res50} \\
    TIFGSM                   &$0.08$ &$0.08$ &$0.26$ &$0.26$ &$0.12$ &$0.12$ \\
    MIFGSM                   &$0.30$ &$0.30$ &$\mathbf{1.16}$ &$1.17$ &$\mathbf{0.68}$ &$0.69$ \\
    AA                       & $0.08$          & $0.14$        & $0.38$          & $0.48$        & $0.16$           & $0.24$         \\
    $\mathbf{GAA}$           & $0.12$               & $0.20$                    & $0.72$ & $1.56$ & $0.44$ & $0.88$ \\
    $\mathbf{PAA_g}$         & $0.14$   & $0.30$        & $0.52$             & $1.27$             & $0.32$          & $0.73$         \\
    $\mathbf{PAA_\ell}$      & $0.12$               & $0.19$                    & $0.34$             & $0.74$             & $0.18$          & $0.49$         \\
    $\mathbf{PAA_{p}}$       & $\mathbf{0.28}$ &$\mathbf{0.36}$& $1.00$ &$\mathbf{1.87}$& $0.56$ & $\mathbf{0.88}$\\ \midrule\midrule
    
    \multirow{2}{*}{} & \multicolumn{2}{c}{Inc-v3$\rightarrow$VGG19} & \multicolumn{2}{c}{Inc-v3$\rightarrow$Den121} & \multicolumn{2}{c}{Inc-v3$\rightarrow$Res50} \\
    TIFGSM                   &$0.16$ &$0.18$ &$0.08$ &$0.09$ &$0.08$ &$0.09$ \\
    MIFGSM                   &$0.56$ &$0.56$ &$0.56$ &$0.57$ &$0.54$ &$0.54$ \\
    AA                       & $0.24$          & $0.66$        & $0.28$          & $1.02$        & $0.24$           & $0.66$         \\
    $\mathbf{GAA}$           & $0.60$ & $2.49$& $0.72$ & $2.38$ & $0.68$ & $2.60$ \\
    $\mathbf{PAA_g}$         & $0.34$             & $1.03$             & $0.38$             & $1.24$             & $0.34$          & $1.24$         \\
    $\mathbf{PAA_\ell}$      & $0.22$          & $0.71$        & $0.32$          & $1.95$        & $0.32$          & $2.13$         \\
    $\mathbf{PAA_{p}}$       & $\mathbf{0.70}$ &$\mathbf{2.55}$& $\mathbf{0.86}$ &$\mathbf{3.37}$& $\mathbf{0.82}$ & $\mathbf{3.10}$\\ \midrule\midrule
    
    \multirow{2}{*}{} & \multicolumn{2}{c}{Res50$\rightarrow$VGG19} & \multicolumn{2}{c}{Res50$\rightarrow$Inc-v3} & \multicolumn{2}{c}{Res50$\rightarrow$Den121} \\
    TIFGSM                 &$0.32$ &$0.33$ &$0.08$ &$0.08$ &$0.44$ &$0.45$ \\
    MIFGSM                  &$2.00$ &$2.02$ &$0.92$ &$0.93$ &$3.96$ &$3.99$ \\
    AA                      & $0.78$          & $2.05$        & $0.54$          & $1.29$        &$1.96$ &$4.93$      \\
    $\mathbf{GAA}$          & $2.14$          & $5.28$        & $0.76$ & $1.78$ & $3.92$ & $9.80$ \\
    $\mathbf{PAA_g}$        & $0.94$          & $3.15$        & $0.44$          & $1.28$        & $1.68$          & $5.96$         \\
    $\mathbf{PAA_\ell}$     & $2.16$ & $5.91$ & $0.62$          & $1.71$        & $3.10$          & $8.54$         \\
    $\mathbf{PAA_{p}}$      & $\mathbf{4.38}$ &$\mathbf{8.46}$& $\mathbf{1.36}$ &$\mathbf{2.48}$& $\mathbf{7.36}$ & $\mathbf{14.88}$\\ \midrule\midrule
    \end{tabular}%
    }
    \caption{Quantitative comparisons with state-of-the-art attacks under the random sample strategy of target label selection. Ours achieve the best performance in most cases.}
    \label{table.optlayerRandom}
\end{table}

\begin{table}[t]
    \centering
    \resizebox{\columnwidth}{!}{
    \begin{tabular}{@{}c|c|ccccc@{}}
    \toprule
\multicolumn{2}{c}{}                        & $2nd$ & $10th$ & $100th$ & $500th$ & $1000th$ \\ \midrule\midrule
$\mathbf{GAA}$      & \multirow{4}{*}{tSuc} &$28.38$       &$14.38$        &$7.6$         &$3.78$         &$0.78$          \\
$\mathbf{PAA_g}$    &                       &$27.18$       &$11.3$        &$4.28$         &$1.62$         &$0.32$          \\
$\mathbf{PAA_\ell}$ &                       &$33.28$       &$15.56$        &$6.86$         &$3.10$         &$0.86$          \\
$\mathbf{PAA_p}$    &                       &$\mathbf{37.98}$       &$\mathbf{21.10}$        &$\mathbf{11.2}$         &$\mathbf{5.12}$         &$\mathbf{1.74}$          \\ \midrule\midrule
$\mathbf{GAA}$      & \multirow{4}{*}{tTR}  &$36.15$       &$19.82$        &$10.75$         &$5.26$         &$1.03$          \\
$\mathbf{PAA_g}$    &                       &$34.94$       &$16.56$        &$6.68$         &$2.64$         &$0.58$          \\
$\mathbf{PAA_\ell}$ &                       &$39.39$       &$19.68$        &$9.32$         &$4.27$         &$1.05$          \\
$\mathbf{PAA_p}$    &                       &$\mathbf{44.90}$       &$\mathbf{26.01}$        &$\mathbf{14.66}$         &$\mathbf{6.74}$         &$\mathbf{2.15}$          \\ \bottomrule
    \end{tabular}
    }
    \caption{Transferability (tSuc and tTR) \wrt $\mathit{2nd}$, $10th$, $100th$, $500th$, and $1000th$ settings. Formally, different target labels lead to different performance and those of lower-ranking lead to worse performance.
    }
    \label{tab:targetselection}
\end{table}

\begin{figure}[t]
    \centering
    \includegraphics[width=0.95\columnwidth,height=8cm]{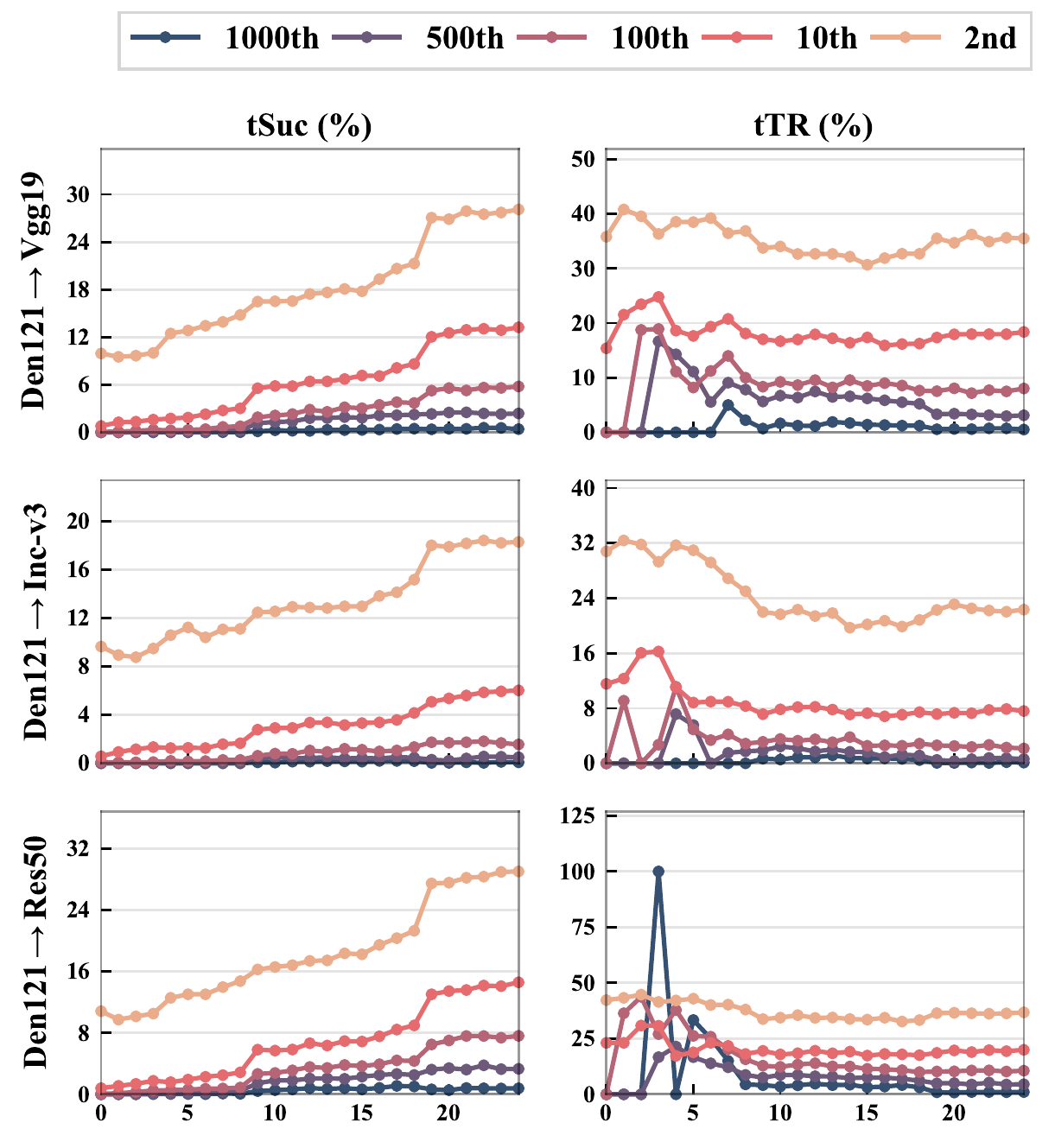}
    \caption{Performance (tSuc and tTR) of $\mathbf{GAA}$ \wrt $\mathit{2nd}$, $10th$, $100th$, $500th$, and $1000th$ settings. Target label of higher ranking leads to better performance.}
    \label{Fig.classGAA}
\end{figure}

\begin{figure*}[t]
    \centering
    \includegraphics[width=\textwidth,height=9.2cm]{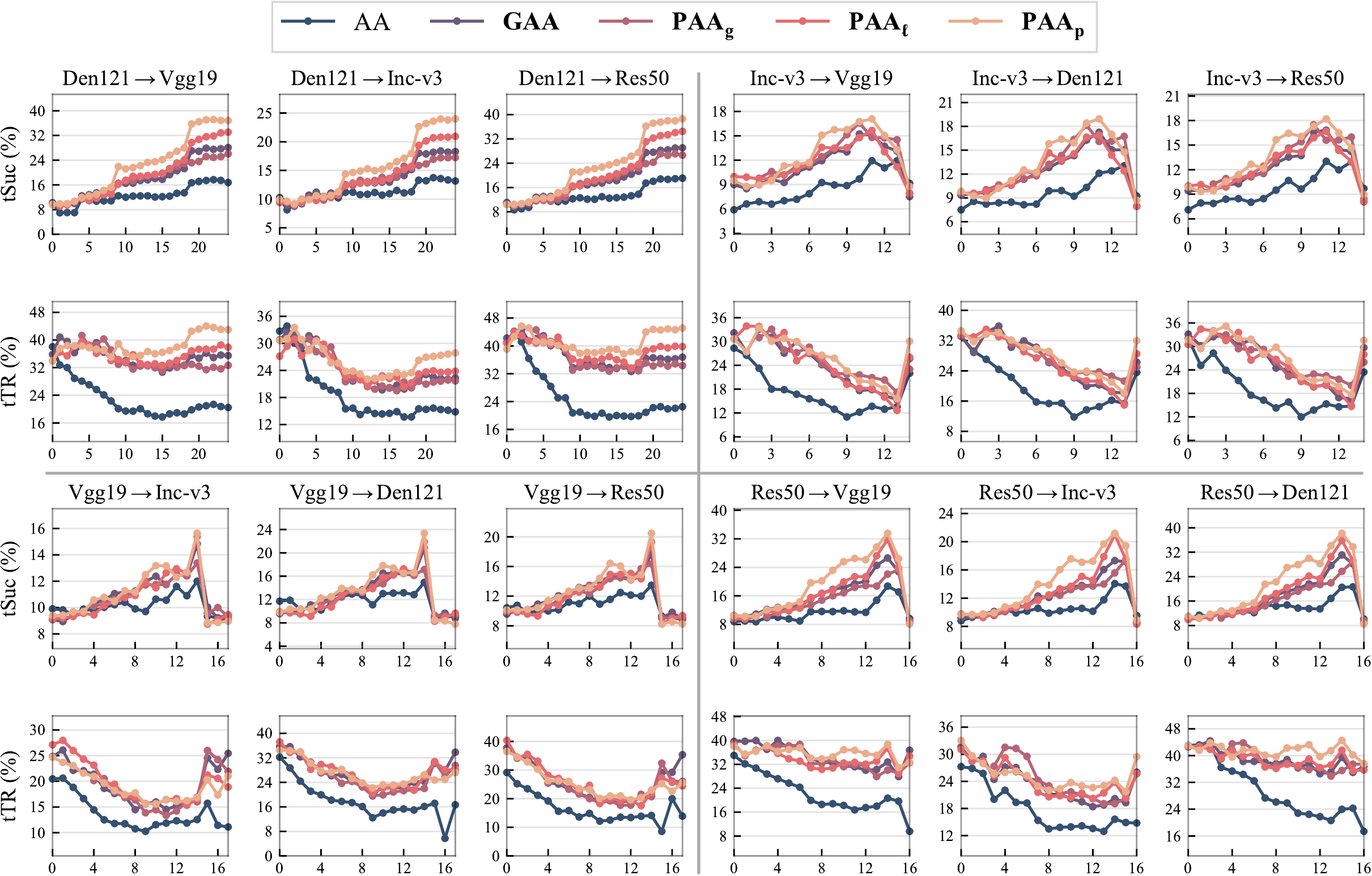}
    \caption{tSuc and tTR performance \wrt relative layer depth for multiple transfer scenarios. The figure is split into four phases: upper left, upper right, bottom left, and bottom right, corresponding to black-box attacks transferring from Den121, Inc-v3, VGG19, and Res50. All of our proposed methods outperform AA in most cases, which indicates the effectiveness of statistic alignment on various layers.}
    \label{Fig.all}
\end{figure*}
\begin{figure}[t]
    \centering
    \includegraphics[width=\columnwidth,height=5.1cm]{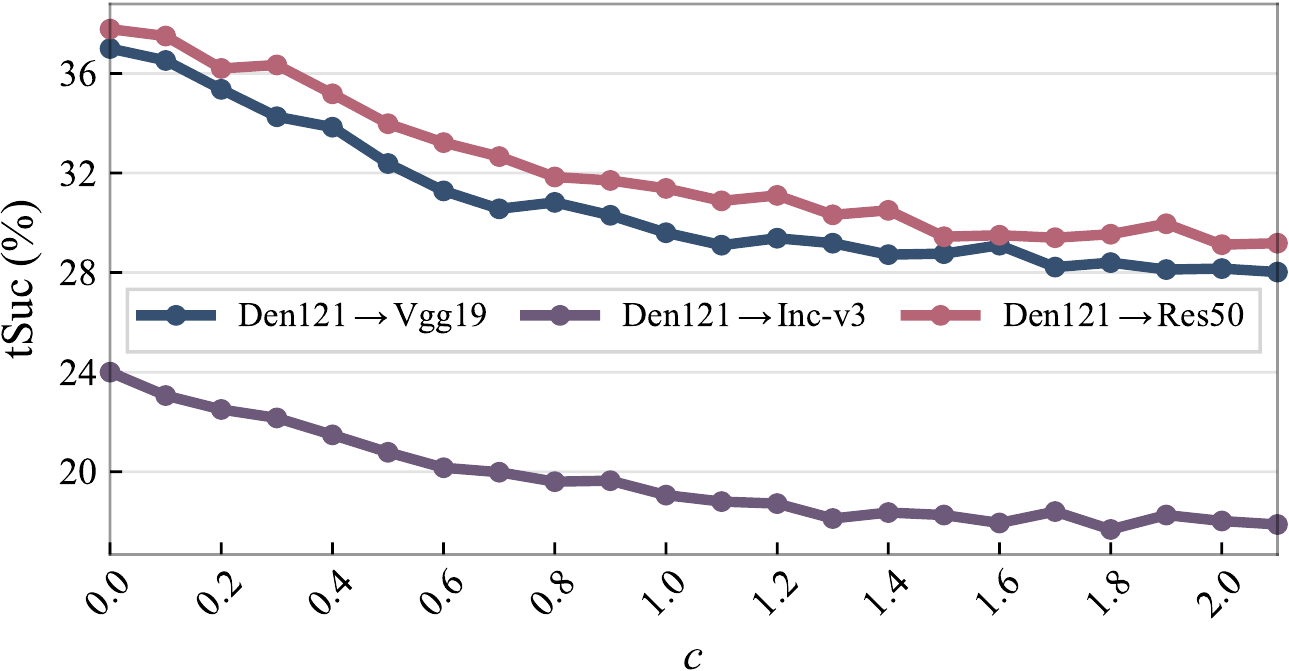}
    \caption{tSuc results \wrt bias $c$ for $\mathbf{PAA_p}$ transferring from Den121 (white-box model) to VGG19, Inc-v3, and Res50 (black-box model). We observe the highest results when $c\!=\!0$, \ie, polynomial with pure second-order terms.}
    \label{Fig.select}
\end{figure}

\section{Experiments}
To make comprehensive comparisons with state-of-the-arts, we conduct a series of experiments to evaluate performance. Specifically, baselines include a feature space targeted attack method: AA~\cite{feature} and two FGSM-based methods: MIFGSM~\cite{mifgsm} and TIFGSM~\cite{tifgsm}. Appendix~\ref{appendix.otherFGSM} give comparisons with other FGSM-based methods.

\paragraph{ImageNet models.} For a better evaluation of transferability, four ImageNet-trained models with different architectures are chosen: VGG-19 with batch-normalization (VGG19)~\cite{vgg}, DenseNet-121 (Den121)~\cite{dense},  ResNet-50 (Res50)~\cite{resnet}, Inception-v3 (Inc-v3)~\cite{inc-v3}.

\paragraph{Dataset.} Attacking images that have already been misclassified is pointless. Hence for each of all 1000 labels in the ImageNet validation set, we randomly select five images (5,000 in total) to perturb, which are correctly classified by all the networks we considered.

\paragraph{Layer decoding scheme.} Following AA, a scheme for layer decoding is employed to present better which layer is chosen for the attack. Generally, layers are arranged from shallow to deep and numbered by relative layer depths, \eg, layer 0 of Res50 (denoted as Res50\textsubscript{[0]}) is near the input layer, and Res50\textsubscript{[16]} is closed to the classification layer. Appendix~\ref{appendix.layer} details the scheme.

\paragraph{Target label selection.} There are two strategies for target label selection: a) random sample adopted in AA. b) choose by ranking. Previous feature space targeted attack methods, \eg, \cite{feature}, gain relatively poor performance. Given the prior knowledge that different target labels involve different transfer difficulties, randomly sampling the target label will lead to fluctuating transfer results (see Appendix~\ref{appendix.fluctuate} for more analysis). For instance, given an image of ``cat", it is easier to fool a model to predict it as a dog than an airplane. To avoid this, we assign $y^{tgt}$ by ranking. For example, $\mathit{2nd}$ indicates that the label of the second high confidence is chosen to be $y^{tgt}$. To give an exhaustive comparison, $2nd$, $10th$, $100th$, $500th$, and $1000th$ settings are adopted. We also report results under the random sample strategy to reveal the raw performance. 

\paragraph{Implementation details.} To make a fair comparison, all methods are set to identical $\ell_\infty$ constraint $\epsilon \!=\! 0.07$, the number of iterations $T \!=\! 20$, and step size $\alpha \!=\! \epsilon/T\!=\!0.0035$. The gallery size is set to $20\times1000$. For $\mathbf{PAA_g}$, we set variance $\sigma^2$ as the mean of squared $\ell_2$ distances of those pairs. For $\mathbf{PAA_p}$, we set bias $c\!=\!0$, and only study the case of power $d=2$. For TIFGSM, we adopt the default kernel length as 15. For MIFGSM, we set the decay factor as $\mu\!=\!1.0$.

\paragraph{Evaluation metrics.} Following AA, we adopt two metrics, \ie, targeted success rate (tSuc) and targeted transfer rate (tTR), to evaluate the transferability of adversarial examples. For tSuc, it equals the percentage of adversarial examples that successfully fool the victim's DNNs. For tTR, 
given an image set that contains adversarial examples that attack the substitute model successfully, tTR is the ratio that how many examples of this set can fool the black-box model too.

\subsection{Comparisons with State-of-the-Art Attacks}
In this section, to comprehensively evaluate adversarial examples' transferability, we firstly attack different white-box models using the random sample strategy for target labels, then transfer the resultant adversarial examples to black-box models. For instance, Den121$\rightarrow$Res50 indicates that we generate adversarial examples from Den121 and transfer them to Res50. Empirically, attack performance varies according to the choice of layers. Under random sample strategy, VGG19\textsubscript{[10]}, Den121\textsubscript{[23]}, Res50\textsubscript{[11]} and Inc-v3\textsubscript{[8]} perform the best, their experimental results are shown in Table~\ref{table.optlayerRandom}.

\subsubsection{Effectiveness of $\mathbf{PAA}$ with Different Kernel Functions}
As demonstrated in Table~\ref{table.optlayerRandom}, all pair-wise alignments show their success in feature space targeted attack. Specifically, comparing with Linear kernel and Gaussian kernel, Polynomial kernel brings the best performance, and our $\mathbf{PAA_p}$ outperforms state-of-the-arts by 6.92\% at most and 1.70\% on average, which shows the effectiveness of our pair-wise alignment. As for the reasons of the performance gains, compared with FGSM-based methods, \ie, TIFGSM and MIFGSM, we exploit the information in the intermediate feature maps to perform highly transferable attacks. Compared with AA, it adopts Euclidean distance for measuring differences so that shows worse performance than ours, demonstrating the effectiveness of our proposed statistic alignment.

\subsubsection{Effectiveness of $\mathbf{GAA}$}
Although $\mathbf{GAA}$ requires quite simple computations to perform attacks, it still shows convincing performance against all black-box models. Specifically, $\mathbf{GAA}$ outperforms the state-of-the-arts by 3.98\% at most and 0.73\% on average, which shows the effectiveness of global alignment between statistics from target and source. Moreover, when choosing Den121 and Res50 as white-box models, it shows comparable performance with $\mathbf{PAA_\ell}$. When it becomes VGG19 or Inc-v3, $\mathbf{GAA}$ achieves the second-best results in most cases. 

\subsection{Ablation Study}
\subsubsection{Transferability \wrt Target Labels}
\label{exp.diffKernel}
Considering different difficulties of target label ${y^{tgt}}$, for $\mathbf{PAA_{p}}$ and $\mathbf{GAA}$, we study how layer-wise transferability varies with "$\mathit{2nd}$", "$\mathit{10th}$", "$\mathit{100th}$", "$\mathit{500th}$", "$\mathit{1000th}$" setup. As illustrated in Figure~\ref{Fig.classPAA} and Figure~\ref{Fig.classGAA}, tSuc and tTR \wrt relative layer depth under above settings are evaluated. Obviously, the independence of layer-wise transferability from different target labels maintains. In other words, different target labels do not affect the layer-wise transferability trends, although further ${y^{tgt}}$ away from ground truth $y$ leads to a more challenging transfer-based attack. For case Den121$\rightarrow$Res50 under $\mathit{2nd}$, we report the results for the optimal layer of Den121 (Den121\textsubscript{[22]}) in Table~\ref{tab:targetselection}. Formally, target labels of different ranks lead to different performance , and the lower-ranking leads to worse performance. Specifically, $2nd$ is the best case, $1000th$ refers to the worst case.

\subsubsection{Transferability \wrt Layers}
In this section, transferability \wrt relative layer depth under $\mathit{2nd}$ is investigated. Involved methods contain $\mathbf{PAA_\ell}$, $\mathbf{PAA_p}$, $\mathbf{PAA_g}$, $\mathbf{GAA}$, and AA. Specifically, given the white-box and black-box model pair, each subfigure of Figure~\ref{Fig.all} illustrates performance under different metric \wrt relative layer depth. As demonstrated in the figure, compared with the Linear kernel, the Polynomial kernel brings about better attack ability on Res50, Inc-v3, and Dense121 white-box. As for the VGG19 white-box, they achieve comparable results. Furthermore, in most of the chosen layers, all of our methods are superior to the baseline AA by a large margin. 

Similar to what is stated in~\cite{feature}, given a white-box model, our layer-wise transferability still holds a similar trend regardless of which black-box models we test. Specifically, for Den121, a deeper layer yields more transferability. For Inc-v3, Vgg19, and Res50, the most powerful attack comes from perturbations generated from optimal middle layers. This phenomenon indicates that adversarial examples generated by our optimal layers can be well transferred to truly unknown models. From the experimental results, under $2nd$, we simply adopt VGG19\textsubscript{[14]}, Den121\textsubscript{[22]}, Res50\textsubscript{[14]}, and Inc-v3\textsubscript{[11]} as our optimal layers. 

\subsubsection{Transferability \wrt Orders}
\label{cccc}
As mentioned above, the Polynomial kernel leads to the most powerful attack. Since larger bias c $(c\geq0)$ results in a greater proportion of lower-order terms in the polynomial, in this section, we study the appropriate value of $c$ under $\mathit{2nd}$ and Den121\textsubscript{[22]} setup. Specifically, we attack Den121 using $\mathbf{PAA_p}$ parameterized by $c$ ranging from 0.0 to 2.0 with a granularity 0.1. As illustrated in Figure~\ref{Fig.select}, from the monotonically decreasing curves, we can achieve the most effective attack when $c=0.0$, where tSuc is 37.00\%, 24.00\%, 37.78\% for VGG19, Inc-v3 , and Res50. Once $c\!=\!1.3$ or larger, tSuc maintains stable. The overall average tSuc for VGG19, Inc-v3, Res50 are 30.78\%, 19.68\%, and 32.12\%.

\section{Conclusion}
In this paper, we propose a novel statistic alignment for feature space targeted attacks. Previous methods utilize Euclidean distance to craft perturbations. However, because of the spatial-related property of this metric, it unreasonably imposes a spatial-consistency constraint on the source and target features. To address this problem, two novel methods, \ie, Pair-wise Alignment Attack and Global-wise Alignment Attack are proposed by employing high-order translation-invariant statistics. Moreover, since randomly selecting target labels results in fluctuating transfer results, we further analyze the layer-wise transferability with different transfer difficulties to obtain highly reliable attacks. Extensive experimental results show the effectiveness of our methods. 


\section*{Acknowledgements}
This work is supported by National Key Research and Development Program of China (No.2018AAA0102200), the National Natural Science Foundation of China (Grant No.61772116, No.61872064, No.62020106008), Sichuan Science and Technology Program (Grant No.2019JDTD0005), The Open Project of Zhejiang Lab (Grant No.2019KD0AB05) and Open Project of Key Laboratory of Artificial Intelligence, Ministry of Education (Grant No.AI2019005).

\bibliographystyle{named}
\bibliography{ijcai21}



\newpage
\appendix
\section*{\centering Appendix}

\begin{table}[!ht]
\centering
\begin{tabular}{@{}ccccc@{}}
\toprule
Layer & Den121 & VGG19 & Inc-v3 & Res50 \\ \midrule\midrule
0     &-1     &256       &C3b1x1        &1       \\
1     &2     &256       &C4a3x3        &2       \\
2     &4        &256       &M        &3       \\
3     &6        &256       &Mix5b        &3,1       \\
4     &6,2        &M       &Mix5c        &3,2       \\
5     &6,4        &512       &Mix5d        &3,3       \\
6     &6,6        &512      &Mix6a        &3,4       \\
7     &6,10        &512       &Mix6b        &3,4,1       \\
8     &6,12        &512       &Mix6c        &3,4,2       \\
9     &6,12,1        &M       &Mix6d        &3,4,3       \\
10    &6,12,3        &512       &Mix6e        &3,4,4       \\
11    &6,12,5        &512       &Mix7a        &3,4,5       \\
12    &6,12,8        &512       &Mix7b        &3,4,6       \\
13    &6,12,10        &512       &Mix7c        &3,4,6,1       \\
14    &6,12,14        &M       &FC        &3,4,6,2       \\
15    &6,12,16        &FC1       &-        &3,4,6,3       \\
16    &6,12,19        &FC2       &-        &3,4,6,3,FC       \\
17    &6,12,22        &FC3       &-        &-       \\
18    &6,12,24        &-       &-        &   -    \\
19    &6,12,24,2        &-       &-        &  -     \\
20    &6,12,24,5        &-       &-        & -      \\
21    &6,12,24,10        &-       &-        & -      \\
22    &6,12,24,12        &-       &-        & -      \\
23    &6,12,24,14        &-       &-        & -      \\
24    &6,12,24,16        &-       &-        &-       \\ \bottomrule
\end{tabular}
\caption{Layer annotations for Den121, VGG19, Inc-v3 and Res50, repectively.}
\label{lookUpTableForLayer}
\end{table}

\begin{figure}[!ht]
    \centering
    \includegraphics[width=\columnwidth]{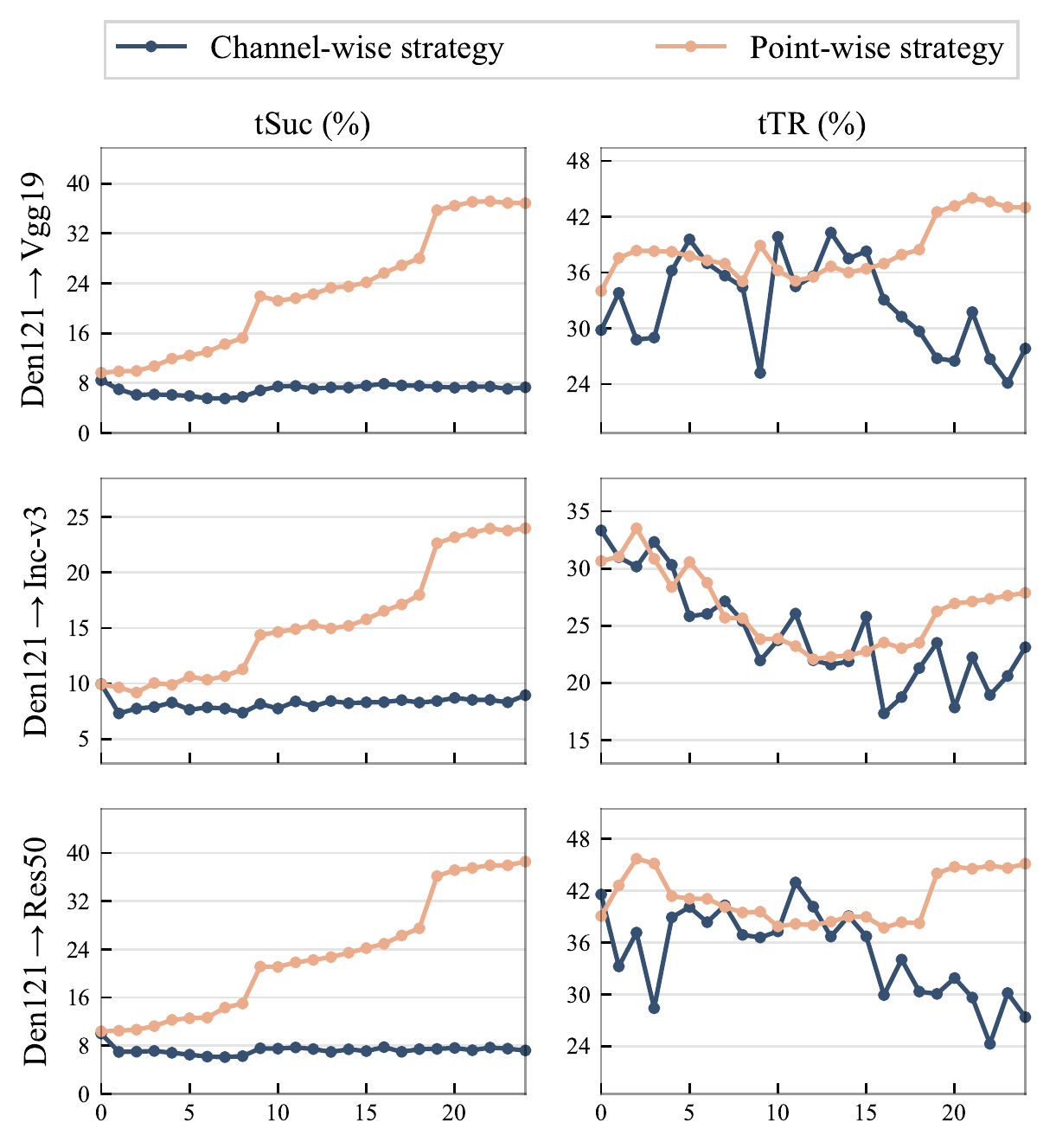}
    \caption{Transfer results under two sample strategies, given Den121 as the white-box model.}
    \label{Fig.sampleStrategy}
\end{figure}

\begin{figure}[!ht]
    \centering
    \includegraphics[width=\columnwidth]{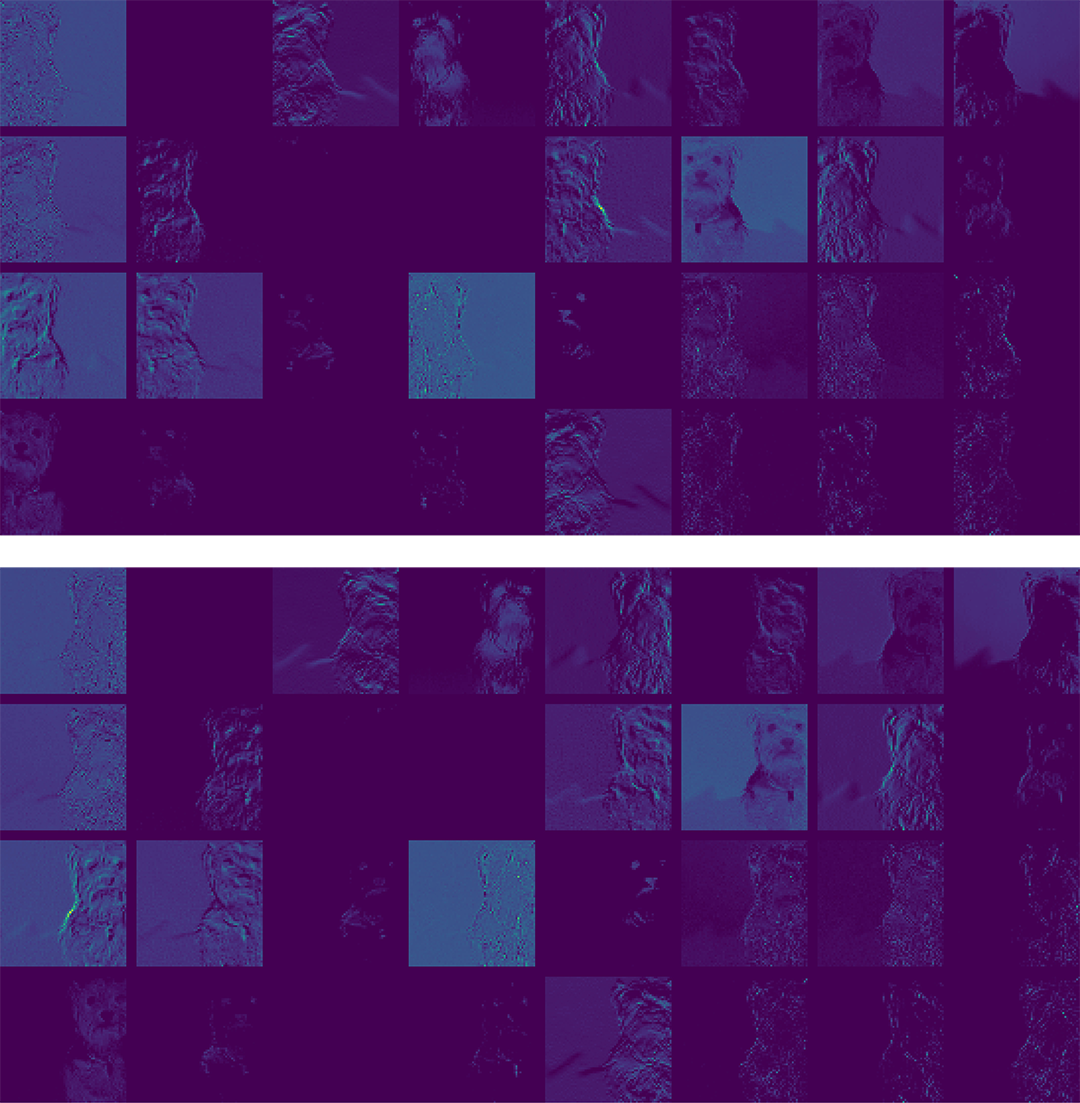}
    \caption{Demonstration of extracted feature maps from the layer $2$ of Inc-v3. Top: a dog on the left. Bottom: same image with horizontal flipped.}
    \label{Fig.demo}
\end{figure}

\section{Layer Decoding}
\label{appendix.layer}
Table~\ref{lookUpTableForLayer} gives the detailed information for the chosen layers of all models we test on. Den121 follows the implementation here: \url{https://github.com/pytorch/vision/blob/master/torchvision/models/densenet.py},Den121 has four denseblocks, each of them has [6,12,24,16] denselayers. \eg, given layer 1, we extract the output of the $2^{nd}$ denselayer of the first denseblock as the feature map, for layer 23, the feature map corresponding to the output of the $14^{nd}$ denselayer of the fourth denseblock. Notably, layer 0 means that we extract the final output before the first block.

The VGG19 model follows the implementation here:\url{https://github.com/pytorch/vision/blob/master/torchvision/models/vgg.py}. The complete layer array for VGG19 is [64, 64, 'M', 128, 128, 'M', 256, 256, 256, 256, 'M', 512, 512, 512, 512, 'M', 512, 512, 512, 512, 'M'], 'M' refers to the MaxPool2d layer, and each number corresponding to how many channels a 3x3 convolutional layer has. FC1, FC2, FC3 refer to the last three linear layers of VGG19. Some layers of the arrays are not chosen since perturbing images based on these layers is helpless. 

The Inc-v3 model follows the implementation here: \url{https://github.com/pytorch/vision/blob/master/torchvision/models/inception.py}. The layer array of Inc-v3 is [C1a3x3, C2a3x3, C2b3x3,M,C3b1x1, C4a3x3, M, Mi5b, Mi5c, Mi5d, Mi6a, Mi6b, Mi6c, Mi6d, Mi6e], 'M' refers to the MaxPool2d layer, others start with 'C' refers to convolutional layers, and the rest represents different inception blocks. 

The Res50 model follows the implementation here: 
\url{ https://github.com/pytorch/vision/blob/master/torchvision/models/resnet.py}. Res50 has 4 layer groups, each layer group has 3, 4, 6, and 3 Bottlenecks, FC is the last linear layer of Res50. Layer 0 means that we choose the output of the first Bottlenecks from $1^{st}$ layer group.

\begin{figure}[t]
     \centering
      \begin{subfigure}[b]{0.24\columnwidth}
         \centering
         \includegraphics[width=\textwidth]{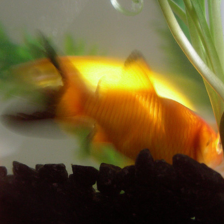}
         \label{fig.ori}
     \end{subfigure}
     \begin{subfigure}[b]{0.24\columnwidth}
         \centering
         \includegraphics[width=\textwidth]{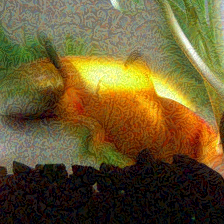}
         \label{fig.example_MIFGSM}
     \end{subfigure}
     \begin{subfigure}[b]{0.24\columnwidth}
         \centering
         \includegraphics[width=\textwidth]{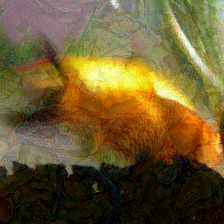}
         \label{fig.example_AA}
     \end{subfigure}
     \begin{subfigure}[b]{0.24\columnwidth}
         \centering
         \includegraphics[width=\textwidth]{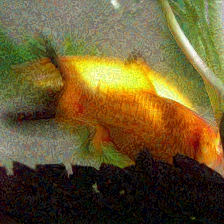}
         \label{fig.example_PAA_p}
     \end{subfigure}
     \vspace{-1.3em}\caption{Visualization of adversarial examples with Den121 as the white-box. Original class: goldfish, targeted class: axolotl. From left to right: Raw, MIFGSM, AA and $\mathbf{PAA_p}$ (both on Den121\textsubscript{[22]}).}
     \label{Fig.visual}
\end{figure}

\section{Visualization of Adversarial Examples}
Please refer to Figure~\ref{Fig.visual}. Obviously, $\mathbf{PAA_p}$ obtains better performance and higher stealthiness than MIFGSM and AA.

\section{Different Sample Strategies}
\label{appendix.sampleStrategy}
There are two strategies to sampling feature maps: pixel-wise and channel-wise. Given source and target feature maps $\vect{S^l}\in\mathbb{R}^{{N_l}\times{M_l}}$, $\vect{T^l}\in\mathbb{R}^{{N_l}\times{M_l}}$, by following different sample strategy, corresponding sample sets $\Omega$ and $Z$ are obtained. Channel-wise strategy means that we sample from the feature maps channel-by-channel (\ie $\Omega\!=\!\{{\vect s_{ i \cdot}}\}_{i=1}^{N_l}$,  $Z\!=\!\{{\vect t_{j \cdot}}\}_{j=1}^{N_l}$). Point-wise strategy changes to pixel-by-pixel (\ie $\Omega\!=\!\{{\vect s_{\cdot i}}\}_{i=1}^{M_l}$,  $Z\!=\!\{{\vect t_{\cdot j}}\}_{j=1}^{M_l}$). Given Dense121 as the white-box model, the transferability results under different strategies are shown in Fig.~\ref{Fig.sampleStrategy}. Obviously, the tSuc results of Channel-wise strategy don't vary with the choices of layers. And for all the layers we choose, Point-wise strategy always outperforms Channel-wise strategy. Thus in our experiments we adopt the Point-wise strategy.

\section{Demonstration of the Extracted Feature Maps from Inc-v3}
\label{appendix.perception}
Convolution kernels do not perform a one-time transformation to produce results from the input. Instead, a small region of input is perceived iteratively, so that feature at every layer still holds a local structure similar to that of the input. As Figure~\ref{Fig.demo} demonstrates. This property of convolution leads to the translation homogeneity of intermediate feature maps. Therefore, measuring only the Euclidean distance between two feature maps will be inaccurate when there are translations, rotations \etc.

\begin{table}[!ht]
    \resizebox{\columnwidth}{!}{%
    \begin{tabular}{@{}ccccccc@{}}
    \toprule
    \multirow{2}{*}{} & \multicolumn{2}{c}{Den121$\rightarrow$VGG19} & \multicolumn{2}{c}{Den121$\rightarrow$Inc-v3} & \multicolumn{2}{c}{Den121$\rightarrow$Res50} \\
                             & tSuc             & tTR            & tSuc             & tTR            & tSuc             & tTR             \\ \midrule\midrule
    AA                       & $17.68$          & $21.36$        & $13.58$          & $15.34$        & $18.86$           & $21.89$         \\
    $\mathbf{GAA}$                       & $27.50$          & $34.92$        & $18.40$          & $22.20$        & $28.38$          & $36.15$         \\
    $\mathbf{PAA_{g}}$                    & $24.92$          & $32.10$        & $17.20$          & $21.62$        & $26.60$          & $34.19$         \\
    $\mathbf{PAA_\ell}$                   &\underline{$31.82$}&\underline{$37.36$}&\underline{$20.80$}&\underline{$23.57$}&\underline{$34.36$}&\underline{$39.42$}\\
    $\mathbf{PAA_{p}}$                     & $\mathbf{37.16}$ &$\mathbf{43.62}$& $\mathbf{23.96}$ &$\mathbf{27.35}$& $\mathbf{37.98}$ & $\mathbf{44.90}$\\ \midrule\midrule
    
    \multirow{2}{*}{} & \multicolumn{2}{c}{VGG19$\rightarrow$Inc-v3} & \multicolumn{2}{c}{VGG19$\rightarrow$Den121} & \multicolumn{2}{c}{VGG19$\rightarrow$Res50} \\
    AA                       & $12.00$          & $12.55$        & $14.94$          & $16.13$        & $13.46$           & $14.13$         \\
    $\mathbf{GAA}$                       & $14.84$          & $16.897$        & $20.86$          & $24.35$        & $17.84$          & $21.00$         \\
    $\mathbf{PAA_{g}}$                    & $13.38$          & $15.99$        & $17.14$          & $22.01$        & $16.34$          & $20.70$         \\
    $\mathbf{PAA_\ell}$                   &\underline{$15.38$}&\underline{$16.52$}&\underline{$21.90$}&\underline{$26.03$}&\underline{$19.3$}&\underline{$22.13$}\\
    $\mathbf{PAA_{p}}$                     & $\mathbf{15.64}$ &$\mathbf{16.83}$& $\mathbf{23.36}$ &$\mathbf{26.48}$& $\mathbf{20.48}$ & $\mathbf{23.03}$\\ \midrule\midrule
    
    \multirow{2}{*}{} & \multicolumn{2}{c}{Inc-v3$\rightarrow$VGG19} & \multicolumn{2}{c}{Inc-v3$\rightarrow$Den121} & \multicolumn{2}{c}{Inc-v3$\rightarrow$Res50} \\
    AA                       & $11.94$          & $13.69$        & $12.1$          & $14.54$        & $13.02$           & $15.31$         \\
    $\mathbf{GAA}$                       & $15.06$          & $17.89$        & $17.26$          & $20.31$        & $16.82$          & $20.31$         \\
    $\mathbf{PAA_{g}}$                    & $14.8$          & $20.91$        & $16.10$          & $23.71$        & $15.60$          & $22.50$         \\
    $\mathbf{PAA_\ell}$                   &\underline{$15.66$}&\underline{$18.15$}&\underline{$16.9$}&\underline{$21.196$}&\underline{$16.68$}&\underline{$20.09$}\\
    $\mathbf{PAA_{p}}$                     & $\mathbf{17.08}$ &$\mathbf{19.74}$& $\mathbf{18.94}$ &$\mathbf{22.75}$& $\mathbf{18.20}$ & $\mathbf{21.58}$\\ \midrule\midrule
    
    \multirow{2}{*}{} & \multicolumn{2}{c}{Res50$\rightarrow$VGG19} & \multicolumn{2}{c}{Res50$\rightarrow$Inc-v3} & \multicolumn{2}{c}{Res50$\rightarrow$Den121} \\
    AA                       & $18.72$          & $20.67$        & $14.06$          & $15.62$        & $20.52$           & $23.96$         \\
    $\mathbf{GAA}$                       & $26.62$          & $32.76$        & $17.32$          & $20.17$        & $31.12$          & $39.45$         \\
    $\mathbf{PAA_{g}}$                    & $22.10$          & $30.03$        & $15.58$          & $19.05$        & $25.74$          & $35.54$         \\
    $\mathbf{PAA_\ell}$                   &\underline{$31.80$}&\underline{$37.45$}&\underline{$21.00$}&\underline{$23.81$}&\underline{$35.88$}&\underline{$41.60$}\\
    $\mathbf{PAA_{p}}$                     & $\mathbf{33.56}$ &$\mathbf{38.70}$& $\mathbf{21.20}$ &$\mathbf{24.24}$& $\mathbf{38.22}$ & $\mathbf{44.52}$\\ \midrule\midrule
    \end{tabular}%
    }
    \caption{Quantitative comparisons with the state-of-the-art attacks. All methods are attacked by choosing the $2nd$ target class. For FGSM-based methods, we report the reproduced results; For Feature space attacks, the best results they ever met are presented. Specifically, optimal layers of four white-boxes are: Den121\textsubscript{[23]}, VGG19\textsubscript{[10]},  Inc-v3\textsubscript{[8]}, Res50\textsubscript{[11]}. Compared with settings with randomly selected target classes, the attack performance is much higher. Nevertheless, Ours still achieves the best performance in all cases, indicating the high reliable of our proposed methods.}
    \label{table.optlayer}
\end{table}

\begin{table}[!ht]
    \resizebox{\columnwidth}{!}{%
    \begin{tabular}{@{}ccccccc@{}}
    \toprule
    \multirow{2}{*}{} & \multicolumn{2}{c}{Res50$\rightarrow$VGG19} & \multicolumn{2}{c}{Res50$\rightarrow$Inc-v3} & \multicolumn{2}{c}{Res50$\rightarrow$Den121} \\
                             & tSuc             & tTR            & tSuc             & tTR            & tSuc             & tTR             \\ \midrule\midrule
    NIFGSM                         & $1.40$     & $1.40$     & $0.50$    & $0.50$    & $3.12$     & $3.12$        \\
    AA-NI                          & $0.72$          & $1.80$        & $0.28$          & $0.64$        & $1.76$     & $4.29$         \\
    $\mathbf{GAA\text{-}NI}$       & $2.04$          & $4.71$        & $0.66$          & $1.53$        & $3.50$     & $9.05$        \\
    $\mathbf{PAA_{p}\text{-}NI}$   & $\mathbf{3.22}$ &$\mathbf{6.46}$& $\mathbf{1.16}$ &$\mathbf{2.31}$& $\mathbf{6.24}$ & $\mathbf{12.7}$\\ \midrule\midrule
    SIFGSM                      & $1.38$     & $1.36$    & $0.64$   & $0.64$        & $3.52$     & $3.52$        \\
    AA-SI                       & $1.32$          & $3.26$        & $0.56$          & $1.12$        & $2.92$           & $6.81$         \\
    $\mathbf{GAA\text{-}SI}$    & $3.08$          & $7.95$        & $0.76$          & $1.86$        & $4.82$     & $13.34$       \\
    $\mathbf{PAA_{p}\text{-}SI}$           & $\mathbf{3.62}$ &$\mathbf{8.04}$ & $\mathbf{1.00}$ &$\mathbf{2.28}$& $\mathbf{6.58}$ & $\mathbf{15.15}$\\ \midrule\midrule
    PIFGSM   & $0.40$          & $0.40$        & $0.42$          & $0.42$        & $1.40$     & $1.41$\\
    AA-PI    & $0.22$  & $1.16$        & $0.30$          & $1.28$        & $0.44$           & $2.21$         \\
    $\mathbf{GAA\text{-}PI}$         & $0.90$  & $2.76$        & $0.54$          & $1.57$        & $1.72$           & $5.99$        \\
    $\mathbf{PAA_{p}\text{-}PI}$     & $\mathbf{1.28}$ &$\mathbf{3.58}$& $\mathbf{0.72}$ &$\mathbf{1.74}$& $\mathbf{2.06}$ & $\mathbf{6.72}$\\ \midrule\midrule
    \end{tabular}%
    }
    \caption{Quantitative comparisons with other FGSM-based methods (NIFGSM, SIFGSM, and PIFGSM). Ours still achieves the best performance in all cases, indicating the highly reliable of our methods.}
    \label{table.comOtherFGSM}
\end{table}

\begin{figure*}[!ht]
    \centering
    \includegraphics[width=\textwidth]{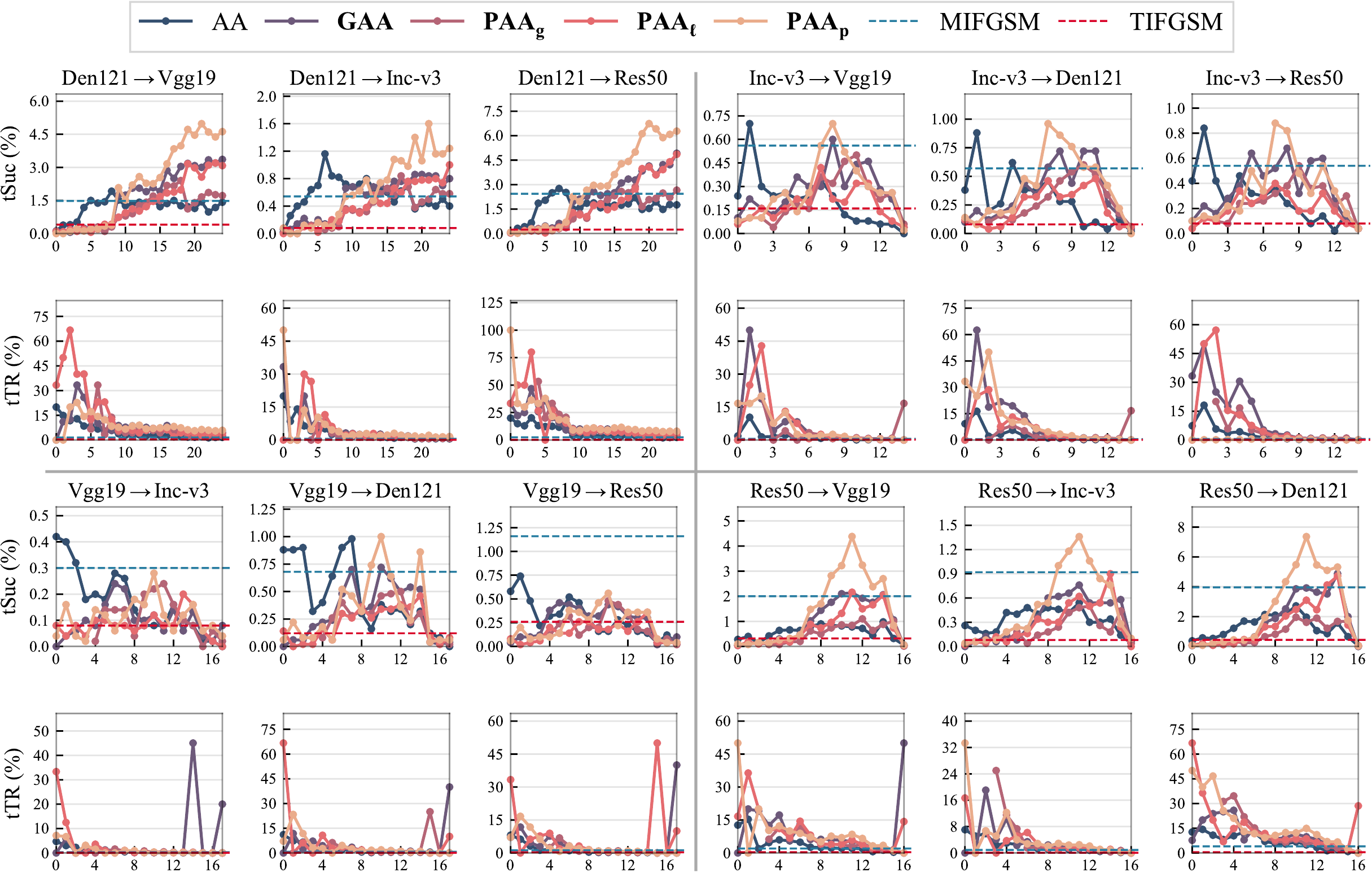}
    \caption{tSuc, and tTR rates \wrt layer depth for multiple transfer scenarios. The figure is split into four phases: upper left, upper right, bottom left, bottom right, corresponding to black-box attacks transferring from Den121, Inc-v3, VGG19, Res50.}
    \label{Fig.randTarClass}
\end{figure*}

\section{The Effectiveness of $\mathbf{PAA}$ and $\mathbf{GAA}$ under 2nd}
\label{appendix.quantiResults}
In this section, under $2nd$, we craft the adversarial perturbations for different white-box models by using the features from their optimal layers, then evaluate the resultant adversarial examples on black-box models. Specifically, optimal layers of four white-boxes are: Den121\textsubscript{[23]}, VGG19\textsubscript{[10]},  Inc-v3\textsubscript{[8]}, Res50\textsubscript{[11]}, and Den121$\rightarrow$Res50 indicates that we firstly generate adversarial examples from Den121 and then transfer them to Res50. As shown in Table~\ref{table.optlayer}, we report the quantitative comparisons with the state-of-the-art attacks. Compared with settings that randomly select target labels, the attack performance is much higher. Nevertheless, Ours still achieves the best performance in all cases, indicating the high reliability of our proposed methods.

\section{Fluctuate Transfer Results under Random Targeted Class Selection}
\label{appendix.fluctuate}
For targeted class selection, there are two strategies: a) random sample, which is adopted in previous methods. b) choose by ranking. Given  the  prior knowledge  that  different targeted class involves different transfer difficulty. For instance, given an image of “cat”, it is easier to fool a model to predict it as a dog than an airplane.  Under this circumstance, the random selection of targeted class will lead to fluctuating transfer results. With the random sample strategy, we study the transferability \wrt relative layer depth. All the experiments are under settings of $\mathit{2nd}$ and $c\!=\!0$. Involved methods containing $\mathbf{PAA_\ell}$, $\mathbf{PAA_p}$, $\mathbf{PAA_g}$, $\mathbf{GAA}$, MIFGSM, TIFGSM, and AA. Given the white-box and black-box model pair, each subfigure of Figure~\ref{Fig.randTarClass} illustrates performances under different metrics \wrt relative layer depth, obviously, it is hard to see the details. The randomness influences the transfer results a lot.

\section{Comparisons with other FGSM-based Methods}
\label{appendix.otherFGSM}
In this section, to further evaluate the effectiveness of our proposed methods ($\mathbf{GAA}$ and $\mathbf{PAA_p}$), given Res50 as the white-box model and using random sample strategy for target label selection, we report the comparisons with other FGSM-based methods, \eg, SIFGSM~\cite{NISI}, NIFGSM~\cite{NISI} and PIFGSM~\cite{pifgsm} in Tabel~\ref{table.comOtherFGSM}. Notably, AA which integrates with MIFGSM is selected as our crucial baseline, and thus the direct FGSM-based competitor is naturally MIFGSM. Only when we modify AA, PAA, and GAA by replacing MIFGSM with other algorithms, \eg, NIFGSM, SIFGSM, and PIFGSM, would it be possible to compare with more state-of-the-arts, otherwise it would be unfair. For simplicity, we denote AA-NI, GAA-NI, and $\mathbf{PAA_p\text{-}NI}$ as the algorithms which replace MIFGSM with NIFGSM in AA, GAA and $\mathbf{PAA_p}$. Specifically, we firstly conduct attacks on the optimal layer of Res50 (Res50\textsubscript{[11]}) using different methods and then transfer the adversarial examples to black-box models, including Den121, Inc-v3, and VGG19. As Tabel~\ref{table.comOtherFGSM} demonstrates, after modifying AA by replacing MIFGSM with NIFGSM and PIFGSM, the new versions of AA achieve lower attack performance compared with original AA. As for SIFGSM, AA-SI obtains a slightly performance gain. Nevertheless, our adversarial examples have better transferability and our $\mathbf{PAA_ps}$ still achieve the most powerful targeted attacks and $\mathbf{GAAs}$ are the second. Specifically, our $\mathbf{PAA_p}$ outperforms the FGSM-based methods by 3.12\% at most and 1.46\%, surpass the AAs by 4.48\% at most and 1.06\% on average, indicating the effectiveness of our statics alignment.

\begin{table}[t]
    \centering
    \resizebox{\columnwidth}{!}{
    \setlength{\tabcolsep}{3.5mm}{
    \begin{tabular}{@{}cccccc@{}}
    \toprule
\multicolumn{1}{c}{}   & $2nd$ & $10th$ & $100th$ & $500th$ & $1000th$ \\ \midrule\midrule
MIFGSM      &$\mathbf{52.78}$       &$\mathbf{23.62}$        &$5.24$         &$0.88$         &$0.22$          \\
AA         &$28.38$       &$14.38$        &$7.6$         &$3.78$         &$0.78$          \\
$\mathbf{GAA}$         &$28.38$       &$14.38$        &$7.6$         &$3.78$         &$0.78$          \\
$\mathbf{PAA_p}$       &$37.98$       &${21.10}$        &$\mathbf{11.2}$         &$\mathbf{5.12}$       &$\mathbf{1.74}$         
\\ \midrule\midrule
NIFGSM      &$\mathbf{50.92}$       &$\mathbf{21.96}$        &$4.70$         &$0.86$         &$0.08$          \\
AA-NI          &$18.50$       &$7.64$        &$2.70$         &$1.24$         &$0.26$          \\
$\mathbf{GAA\text{-}NI}$         &$28.60$       &$14.02$        &$7.22$         &$3.40$         &$0.86$          \\
$\mathbf{PAA_p\text{-}{NI}}$       &${37.20}$       &${20.98}$        &$\mathbf{10.70}$         &$\mathbf{4.94}$         &$\mathbf{1.74}$         
\\ \midrule\midrule
SIFGSM     &$\mathbf{52.20}$       &$\mathbf{20.44}$        &$5.28$         &$1.16$         &$0.22$          \\
AA-SI          &$22.94$       &$9.46$        &$4.30$         &$2.42$         &$0.78$          \\
$\mathbf{GAA\text{-}SI}$         &$29.88$       &$15.1$        &$9.06$         &$4.32$         &$0.90$          \\
$\mathbf{PAA_p\text{-}{SI}}$       &${37.12}$       &${19.24}$        &$\mathbf{11.16}$         &$\mathbf{5.80}$         &$\mathbf{1.62}$          
\\ \midrule\midrule
PIFGSM      &$\mathbf{36.22}$       &$\mathbf{12.76}$        &$2.88$         &$0.50$         &$0.08$          \\
AA-PI          &$12.32$       &$4.14$        &$1.22$         &$0.54$         &$0.16$          \\
$\mathbf{GAA\text{-}PI}$         &$20.80$       &$8.46$        &$3.18$         &$1.08$         &$0.16$          \\
$\mathbf{PAA_p\text{-}{PI}}$       &${27.64}$       &${11.48}$        &$\mathbf{4.70}$         &$\mathbf{1.46}$         &$\mathbf{0.26}$          
\\ \midrule
\midrule
 \end{tabular}}
}
\caption{Transferability (tSuc) \wrt $\mathit{2nd}$, $10th$, $100th$, $500th$, and $1000th$ settings for other FGSM-based methods (\eg, NIFGSM, SIFGSM and PIFGSM). Our $\mathbf{GAA}$ and $\mathbf{PAA_ps}$ obtain noticeable performance gains compared with AA for all cases. $\mathbf{PAA_ps}$ achieve comparable performance to FGSM-based methods under $10th$, and outperforms them by a large margin under $100th$, $500th$, and $1000th$ setting.
}
\label{tab:tSuc-otherFGSM}
\end{table}

\begin{table}[t]
    \centering
    \resizebox{\columnwidth}{!}{
    \setlength{\tabcolsep}{3.5mm}{
    \begin{tabular}{@{}cccccc@{}}
    \toprule
\multicolumn{1}{c}{}   & $2nd$ & $10th$ & $100th$ & $500th$ & $1000th$ \\ \midrule\midrule
MIFGSM      &$\mathbf{52.78}$       &$23.62$        &$5.24$         &$0.88$         &$0.22$          \\
AA          &$34.94$       &$16.56$        &$6.68$         &$2.64$         &$0.58$          \\
$\mathbf{GAA}$         &$36.15$       &$19.82$        &$10.75$         &$5.26$         &$1.03$          \\
$\mathbf{PAA_p}$       &${44.90}$       &$\mathbf{26.01}$        &$\mathbf{14.66}$         &$\mathbf{6.74}$         &$\mathbf{2.15}$          
\\ \midrule\midrule
NIFGSM      &$\mathbf{51.12}$       &$22.05$        &$4.72$         &$0.86$         &$0.08$          \\
AA-NI          &$21.95$       &$10.04$        &$3.99$         &$1.64$         &$0.33$          \\
$\mathbf{GAA\text{-}NI}$         &$36.34$       &$19.26$        &$10.12$         &$4.70$         &$1.14$          \\
$\mathbf{PAA_p\text{-}NI}$       &${44.49}$       &$\mathbf{25.84}$        &$\mathbf{13.94}$         &$\mathbf{6.55}$         &$\mathbf{2.18}$         
\\ \midrule\midrule
SIFGSM     &$\mathbf{52.29}$       &$20.47$        &$5.29$         &$1.16$         &$0.22$          \\
AA-SI          &$28.19$       &$12.82$        &$6.36$         &$3.47$         &$1.13$          \\
$\mathbf{GAA\text{-}SI}$         &$38.80$       &$21.65$        &$13.39$         &$6.28$         &$1.30$          \\
$\mathbf{PAA_p\text{-}SI}$       &${43.78}$       &$\mathbf{23.93}$        &$\mathbf{14.67}$         &$\mathbf{7.87}$         &$\mathbf{2.07}$          
\\ \midrule\midrule
PIFGSM      &$\mathbf{36.37}$       &$12.81$        &$2.89$         &$0.50$         &$0.08$          \\
AA-PI          &$15.92$       &$6.35$        &$2.06$         &$1.02$         &$0.31$          \\
$\mathbf{GAA\text{-}PI}$         &$28.10$       &$12.31$        &$4.87$         &$1.68$         &$0.26$          \\
$\mathbf{PAA_p\text{-}PI}$       &${33.81}$       &$\mathbf{14.96}$        &$\mathbf{6.51}$         &$\mathbf{2.05}$         &$\mathbf{0.34}$          
\\ \midrule\midrule
    \end{tabular}}
    }
    \caption{Transferability (tTR) \wrt $\mathit{2nd}$, $10th$, $100th$, $500th$, and $1000th$ settings for other FGSM-based methods (\eg, NIFGSM, SIFGSM and PIFGSM). Our $\mathbf{GAA}$ and $\mathbf{PAA_ps}$ obtain noticeable performance gains compared with AA for all cases. $\mathbf{PAA_ps}$ surpassing all FGSM-based methods by a large margin under $10th$, $100th$, $500th$, and $1000th$.
    }
    \label{tab:tTR-otherFGSM}
\end{table}

\section{Transferability \wrt Target Labels on other FGSM-based Methods}
Follow the setting in Appendix~\ref{appendix.otherFGSM} except changing the white-box model to Den121, we report the results (tSuc and tTR) on black-box model Res50 under $2nd$, $10th$, $100th$, $500th$, and $1000th$ for the optimal layer of Den121 (Den121$\textsubscript{[23}$) in Table~\ref{tab:tSuc-otherFGSM} and Tabel~\ref{tab:tTR-otherFGSM}. As demonstrated in the tables, formally, lower ranking leads to worse performance. Intuitively, the distributions from
target images of lower-ranking class are more different from that from source images, which leads to a larger discrepancy to align as well as lower performance. Compared with the feature space targeted attack methods AA, our methods obtain noticeable performance gains on both tSuc and tTR, which shows the effectiveness or the proposed statistic alignment. Specifically, our $\mathbf{PAA_ps}$ obtain the best performance, and $\mathbf{GAA}$ takes second place. Compared with FGSM-based methods, for tSuc, our $\mathbf{PAA_ps}$ achieve comparable performance to FGSM-based methods under $10th$, and outperforms them by a large margin under $100th$, $500th$, and $1000th$ setting. For tTR, $\mathbf{PAA_ps}$ surpassing all FGSM-based methods by a large margin under $10th$, $100th$, $500th$, and $1000th$. It is worth noting that FGSM-based methods perform better than ours under $2nd$ setting. However, since $2nd$ refers to the easiest transfer difficulty, it is not applicable in real world compared with other setting, \ie, $10th$, $100th$, $500th$, and $1000th$. Investigating this phenomenon needs further theoretical studies, which is not our main purpose in this paper.

\end{document}